\documentclass[10pt,twocolumn,letterpaper]{article}

\usepackage{cvpr}
\usepackage{times}
\usepackage{epsfig}
\usepackage{graphicx}
\usepackage{amsmath}
\usepackage{amssymb}

\usepackage{latexsym}
\usepackage{url}

\usepackage{multirow}
\usepackage{enumitem}
\usepackage{algorithm}
\usepackage{stfloats}
\usepackage{booktabs}
\usepackage{colortbl}
\usepackage{subcaption}

\usepackage{array}   
\usepackage{graphics}
\usepackage{arydshln}

\usepackage{calc}
\usepackage{comment}

\usepackage{color, colortbl}
\definecolor{Gray}{gray}{0.93}

\newlength\myheight
\newlength\mydepth
\settototalheight\myheight{Xygp}
\newcommand{\fullname}{\textsc{\textbf{Pre}-trained \textbf{v}ision-\textbf{a}nd-\textbf{l}anguage bas\textbf{e}d \textbf{n}aviga\textbf{t}or }}
\newcommand{\short}{\textsc{Prevalent}}

\usepackage{wrapfig}

\newcommand{\RN}[1]{%
	\textup{\lowercase\expandafter{\it \romannumeral#1}}%
}

\newcommand{\beq}{\vspace{0mm}\begin{equation}}
\newcommand{\eeq}{\vspace{0mm}\end{equation}}
\newcommand{\beqs}{\vspace{0mm}\begin{eqnarray}}
\newcommand{\eeqs}{\vspace{0mm}\end{eqnarray}}
\newcommand{\barr}{\begin{array}}
\newcommand{\earr}{\end{array}}

\newcommand{\Amat}[0]{{{\bf A}}}

\newcommand{\Hmat}{{\bf H}}

\newcommand{\Kmat}[0]{{{\bf K}}\xspace}

\newcommand{\Mmat}{{\bf M}}

\newcommand{\Qmat}[0]{{{\bf Q}}\xspace}

\newcommand{\Vmat}[0]{{{\bf V}}}
\newcommand{\Wmat}[0]{{{\bf W}}}

\newcommand{\av}{{\boldsymbol{a}}}
\newcommand{\bv}[0]{{\boldsymbol{b}}}

\newcommand{\hv}[0]{{\boldsymbol{h}}}

\newcommand{\sv}{{\boldsymbol{s}}}

\newcommand{\xv}{\boldsymbol{x}}

\newcommand{\zv}{\boldsymbol{z}}

\newcommand{\thetav}{\boldsymbol{\theta}}

\newcommand{\tauv}[0]{{\boldsymbol{\tau}}}

\newcommand{\R}{\mathbb{R}}
\newcommand{\E}{\mathbb{E}}

\newcommand{\Xcal}{\mathcal{X}}

\newcommand{\Lcal}{\mathcal{L}}

\newcommand{\Acal}{\mathcal{A}}

\newcommand{\Dcal}{\mathcal{D}}

\newcommand{\Tcal}{\mathcal{T}}
\newcommand{\Mcal}{\mathcal{M}}

\newcommand{\Scal}{\mathcal{S}}

\DeclareMathOperator{\RR}{\mathbb{R}} 

\usepackage[pagebackref=true,breaklinks=true,letterpaper=true,colorlinks,bookmarks=false]{hyperref}

\cvprfinalcopy 

\def\cvprPaperID{8620} 

\ifcvprfinal\fi
\begin{document}

\title{Towards Learning a Generic Agent for\\Vision-and-Language Navigation via Pre-training}

\author{
Weituo Hao$^{1 \dagger  \ddagger} $,~~ Chunyuan Li$^{2 \dagger} $\thanks{Corresponding author~~ $^{ \dagger}$Equal Contribution \hspace{1mm} $^{ \ddagger}$Work performed during an internship at MSR}, ~~Xiujun Li$^{2}$, ~~Lawrence Carin$^{1}$,  ~~Jianfeng Gao$^{2}$ \\ $^{1}$Duke University ~~~ $^{2}$Microsoft Research, Redmond \\
{\tt\small \{weituo.hao, lcarin\}@duke.edu~~~ \{chunyl,xiul,jfgao\}@microsoft.com}
}

\maketitle

\begin{abstract}
Learning to navigate in a visual environment following natural-language instructions is a challenging task, because the multimodal inputs to the agent are highly variable, and the training data on a new task is often limited. We present the first pre-training and fine-tuning paradigm for vision-and-language navigation (VLN) tasks. By training on a large amount of image-text-action triplets in a self-supervised learning manner, the pre-trained model provides generic representations of visual environments and language instructions. It can be easily used as a drop-in for existing VLN frameworks, leading to the proposed agent \short{}\footnote{\fullname}. It learns more effectively in new tasks and generalizes better in a previously unseen environment. The performance is validated on three VLN tasks. On the Room-to-Room~\cite{anderson2018vision} benchmark, our model improves the state-of-the-art from 47\% to 51\% on success rate weighted by path length. Further, the learned representation is transferable to other VLN tasks. On two recent tasks, vision-and-dialog navigation~\cite{thomason2019vision} and ``Help, Anna!''~\cite{nguyen2019help}, the proposed \short{} leads to significant improvement over existing methods, achieving a new state of the art.
\end{abstract}

\section{Introduction}
Learning to navigate in a photorealistic home environment based on natural language instructions has attracted increasing research interest ~\cite{savva2017minos,kolve2017ai2,das2018embodied,anderson2018vision,chen2019touchdown}, as it provides insight into core scientific questions about multimodal representations. It also takes a step toward real-world applications, such as personal assistants and in-home robots.
Vision-and-language navigation (VLN) presents a challenging reasoning problem for agents, as the multimodal inputs are highly variable, inherently ambiguous, and often under-specified. 


Most previous methods build on the sequence-to-sequence architecture~\cite{sutskever2014sequence}, where the instruction is encoded as a sequence of words, and the navigation trajectory is decoded as a sequence of actions, enhanced with attention mechanisms~\cite{anderson2018vision,wang2018reinforced,ma2019self} and beam search~\cite{fried2018speaker}.  While a number of methods~\cite{misra2017mapping,monroe2017colors,wang2018look} have been proposed to improve language understanding, common to all existing work is that the agent learns to understand each instruction from scratch or in isolation, without collectively leveraging prior vision-grounded domain knowledge.

However, each instruction in practice only loosely aligns with the desired navigation path, making it imperfect for the existing paradigm of learning to understand the instruction from scratch. This is because 
$(\RN{1})$ every instruction only partially characterizes the trajectory. It can be ambiguous to interpret the instructions, without grounding on the visual states. 
$(\RN{2})$ The objects in visual states and language instructions may share various common forms/relationships, and therefore it is natural to build an informative joint representation beforehand, and use this ``common knowleldge'' for transfer learning in downstream tasks.

\begin{figure}[t!]
	\vspace{-0mm}\centering
	\includegraphics[width=6.75cm]{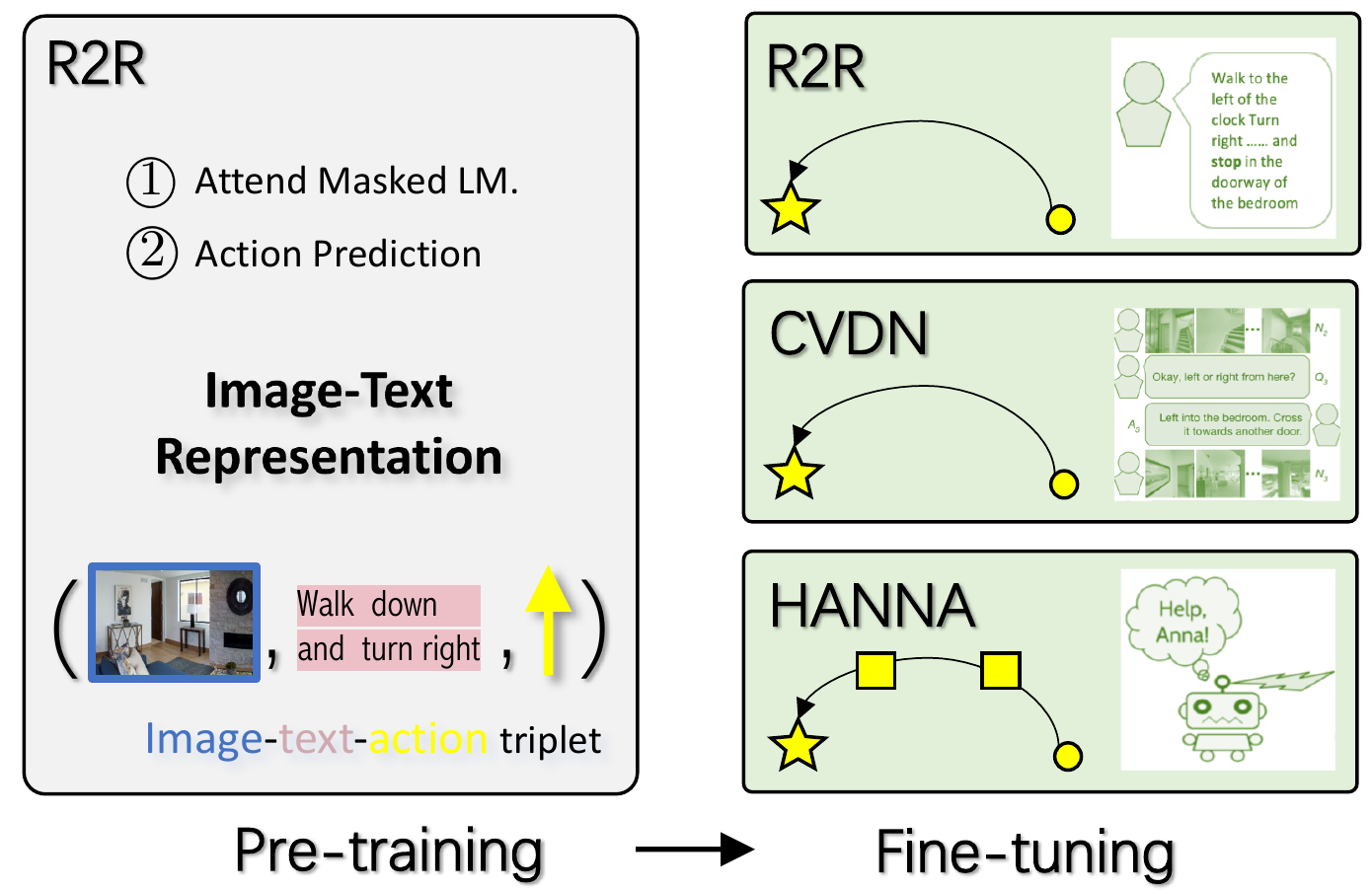}  
	\vspace{-0mm}
	\caption{Illustration of the proposed pre-training and fine-tuning paradigm for VLN. The image-text-action triplets are collected from the R2R dataset. The model is pre-trained with two self-supervised learning objectives, and fine-tuned for three tasks: R2R, CVND and HANNA. R2R is an in-domain task, where the language instruction is given at the beginning, describing the full navigation path. CVND and HANNA are out-of-domain tasks; the former is to navigate based on dialog history, while the latter is an interactive environment, where intermediate instructions are given in the middle of navigation.}
	\vspace{-3mm}
	\label{fig:pretrainig_finetuning_scheme}
\end{figure}

To address this natural ambiguity of instructions more effectively, we propose to pre-train an encoder to align language instructions and visual states for joint representations. The image-text-action triplets at each time step are independently fed into the model, which is trained to predict the masked word tokens and next actions, thus formulating the VLN pre-training in the self-learning paradigm. The complexity of VLN learning can then be reduced by eliminating language understandings which lack consensus from visual states. The pre-trained model plays the role of providing generic image-text representations, and is applicable to most existing approaches to VLN, leading to our agent \short{}. We consider three VLN scenarios as downstream tasks: Room-to-room (R2R)~\cite{anderson2018vision}, cooperative vision-and-dialog navigation (CVDN)~\cite{thomason2019vision}, and ``Help, Anna!'' (HANNA)~\cite{nguyen2019help}. The overall pre-training and finetuning pipeline is shown in Figure~\ref{fig:pretrainig_finetuning_scheme}.

Comprehensive experiments demonstrate strong empirical performance of \short{}. The proposed \short{} achieves a new state of the art on all three tasks~\footnote{Among \emph{all} public results at the time of this submission.}.
Comparing with existing methods, it adapts faster, and generalizes better to unseen environments and new tasks.
Our code and pre-trained model is released on GitHub~\footnote{\url{https://github.com/weituo12321/PREVALENT}}.

\section{Related Work}

\paragraph{Vision-language pre-training}
Vision-Language Pre-trainig (VLP) is a rapidly growing research area. The existing approaches employ BERT-like objectives~\cite{devlin2018bert} to learn cross-modal representation for various vision-language problems, such as visual question-answering, image-text retrieval and image captioning~\etc~\cite{sun2019videobert,tan2019lxmert,lu2019vilbert,zhou2019unified,su2019vl,li2019unicoder}.
However, these VLP works focus on learning representations only for vision-language domains. This paper presents the first pre-trained models, grounding vision-language understanding with actions in a reinforcement learning setting. Further, existing VLP methods require faster R-CNN features as visual inputs~\cite{girshick2015fast,anderson2018bottom}, which are not readily applicable to VLN. State-of-the-art VLN systems are based on panoramic views (\eg, 36 images per view for R2R), and therefore it is computationally infeasible to extract region features for all views and feed them into the agent.    

\paragraph{Vision-and-language navigation} Various methods have been proposed for learning to navigate based on vision-language cues. In \cite{fried2018speaker} a panoramic action space and a ``speaker" model were introduced for data augmentation. 
A novel neural decoding scheme was proposed in \cite{ke2019tactical} with search, to balance global and local information. 
To improve the alignment of the instruction and visual scenes, a visual-textual co-grounding attention mechanism was proposed in~\cite{ma2019self}, which is further improved with a progress monitor~\cite{ma2019regretful}. To improve the generalization of the learned policy to unseen environments, reinforcement learning has been considered, including planning~\cite{wang2018look}, and exploration of unseen environments using a off-policy method~\cite{wang2018reinforced}. An environment dropout was proposed \cite{tan2019learning} to generate more environments based on the limited data, so that it can generalize well to unseen environments. These methods are specifically designed for particular tasks, and hard to generalize for new tasks. In this paper, we propose the first generic agent that is pre-trained to effectively understand vision-language inputs for a broad range of navigation tasks, and can quickly adapt to new tasks. The most related agent to ours is \textsc{PreSS}~\cite{li2019robust}. However, our work is different from ~\cite{li2019robust} from two perspectives: 
$(\RN{1})$ \textsc{PreSS} employs an off-the-shelf BERT~\cite{devlin2018bert} model for language instruction understanding, while we pre-train a vision-language encoder from scratch, specifically for the navigation tasks.
$(\RN{2})$ \textsc{PreSS} only focuses on the R2R task, while we verify the effectiveness of our pre-trained model on three tasks, including two out-of-domain navigation tasks.

\vspace{-1mm}
\section{Background}
The VLN task can be formulated as a Partially Observable Markov Decision Process (POMDP) $\Mcal = \left \langle \Scal, \Acal, P_s, r \right \rangle$, where $\Scal$ is the visual state space, $\Acal$ is a discrete action space, $P_s$ is the unknown environment distribution from which we draw the next state, and $r \in \RR$ is the reward function. At each time step $t$, the agent first observes an RGB image $\sv_t \in \Scal$, and then takes an action $\av_t \in \Acal$. This leads the simulator to generate a new image observation $\sv_{t+1} \sim P_s(\cdot|\sv_t,\av_t)$ as the next state. 
The agent interacts with the environment sequentially, and generates a trajectory of length $T$. 
The episode ends when the agent selects the special $\mathtt{STOP}$ action, or when a pre-defined maximum trajectory length is reached. The navigation is successfully completed if the trajectory $\tauv$ terminates at the intended target location. 

In a typical VLN setting, the instructions are represented as a set $\Xcal = \{\xv_i\}_{i=1}^M$, where $M$ is the number of alternative instructions, and each instruction $\xv_i$ consists of a sequence of $L_i$ word tokens, $\xv_i = [x_{i,1}, x_{i,2}, ..., x_{i,L_i}]$. 
The training dataset $\Dcal_{E} = \{\tauv, \xv \}$ consists of pairs of the instruction $\xv$ together with its corresponding expert trajectory $\tauv$.
%
The agent then learns to navigate via performing maximum likelihood estimation (MLE) of the policy $\pi$, based on the individual sequences: 
\begin{align} 
\hspace{-5mm}
\vspace{-5mm}
 \max_{\thetav}~ \Lcal_{\thetav}(\tauv, \xv) 
\! = \! \log \pi_{\thetav}(\tauv|\xv)
\! = \!\sum_{t=1}^{T}  \log \pi_{\thetav}(\av_t | \sv_t, \xv), 
\label{eq_iid}
\end{align}
where $\thetav$ are the policy parameters. 
The policy is usually parameterized as an attention-based Seq2Seq model~\cite{anderson2018vision,fried2018speaker}, trained in the teacher-forcing fashion, \ie, the ground-truth states $\sv_t$ are provided at every step in training. 
This allows reparameterization of the policy as an encoder-decoder architecture, by considering a function decomposition $\pi_{\thetav} = f_{\thetav_{E}} \circ  f_{\thetav_{D}}$:
\begin{itemize}
    \item  A {\em vision-language encoder} $f_{\thetav_{E}}: \{\sv_t, \xv\} \rightarrow  \zv_t$, where a joint representation $\zv_t$ at time step $t$ is learned over the visual state $\sv_t$ and the language instruction $\xv$.
    \vspace{-4mm}
    \item An {\em action decoder} $f_{\thetav_{D}}: \{\sv_t, \zv_t\} \rightarrow  \av_t$. 
    For each joint representation $\sv_t$, we ground it with $\sv_t$ via neural attention, and decode into actions $\av_t$.
    \vspace{-2mm}
\end{itemize}

Successful navigation largely depends on precise joint understanding of natural language instructions and the visual states \cite{Thomason:19}. We isolate the encoder stage, and focus on pre-training a generic vision-language encoder for various navigation tasks. 



\section{Pre-training Models}
Our pre-training model aims to provide joint representations for image-text inputs in VLN.
\subsection{Input Embeddings}
The input embedding layers convert the inputs (\ie, panoramic views and language instruction) into two sequences of features:  image-level visual embeddings and word-level sentence embeddings.

\begin{figure}[t!]
	\vspace{-0mm}\centering
	\begin{tabular}{cc}
	    \hspace{-3mm}		
		\includegraphics[height=5.4cm]{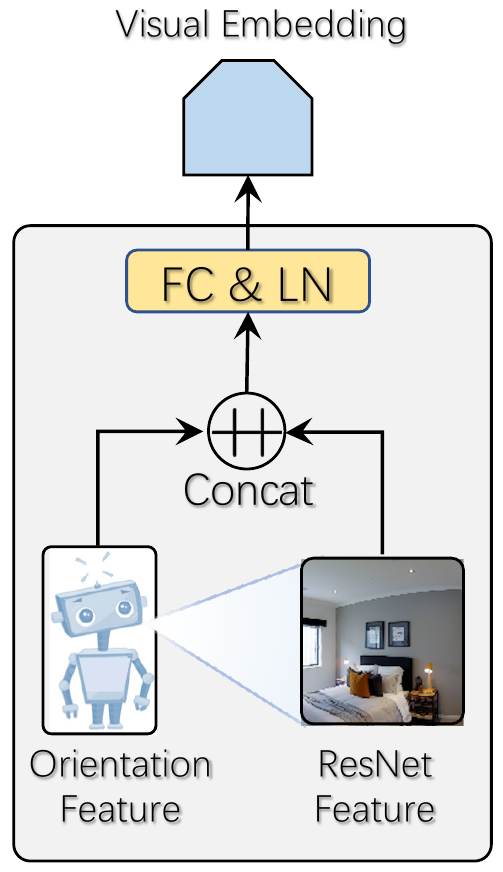} &
		 \hspace{3mm}		
		\includegraphics[height=5.4cm]{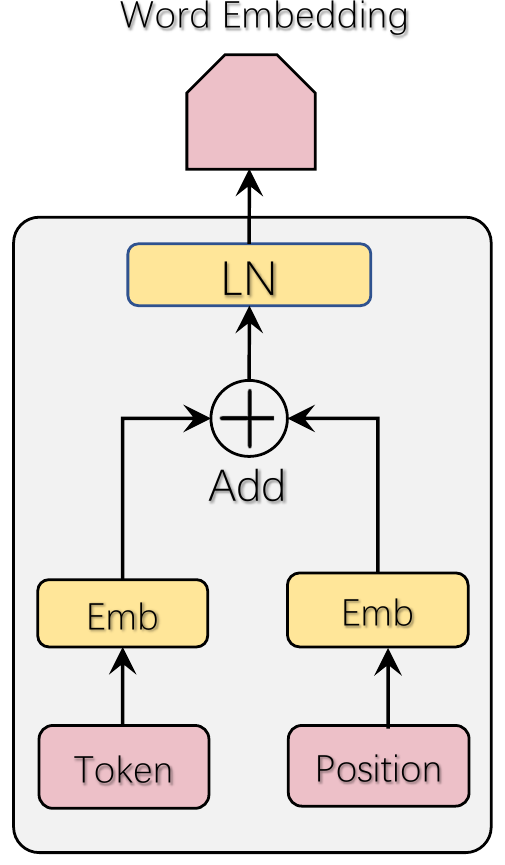}\\
		(a) Visual embedding &
		\hspace{3mm}		
		(b) Text embedding 
	\end{tabular}
	\vspace{-0mm}
	\caption{Illustration for the representation procedure of (a) visual embedding and (b) text embedding. $\mathtt{FC}$ is the fully-connected layer, and $\mathtt{LN}$ is the layer-normalization layer.}
	\vspace{-0mm}
	\label{fig:vision_lang_emb}
\end{figure}

\paragraph{Visual Embedding} Following~\cite{fried2018speaker}, we employ panoramic views as visual inputs to the agent. Each panoramic view consists of 36 images in total (12 angles, and 3 camera poses per angle): $\sv = [s_1, \cdots, s_{36}]$. Each image is represented as a 2176-dimensional feature vector $s = [s_{v},  s_{p}]$, as a result of the concatenation of two vectors: 
$(\RN{1})$ The 2048-dimensional visual feature $s_{v}$ output by a Residual Network (ResNet) of the image~\cite{he2016deep}; 
$(\RN{2})$ the 128-dimensional orientation feature vector $s_{p}$ that repeats $[\sin \psi ; \cos \psi ; \sin \omega; \cos \omega ]$ 32 times, where $\psi$ and $\omega$ are the heading and elevation poses, respectively~\cite{fried2018speaker}. The embedding for each image is:  
\begin{align} 
\hspace{-0mm}
\hv = \text{Layer-Norm}(  \Wmat_e s + \bv_e) ) 
\label{eq_visual_emb}
\end{align}
where $ \Wmat_e \in \R^{d_h \times 2176}$ is a weight matrix, and $\bv_e \in \R^{d_h}$ is the bias term; $d_h=768$ in our experiments. Layer normalization (LN)~\cite{ba2016layer} is used on the output of this fully connected (FC) layer. An illustration of the visual embedding is shown in Figure~\ref{fig:vision_lang_emb}(a).

\paragraph{Text Embedding} The embedding layer for the language instruction follows the standard Transformer, where LN is applied to the summation of the token embedding and position embedding.  An illustration of the text embedding is shown in Figure~\ref{fig:vision_lang_emb}(b).

\subsection{Encoder Architecture}

Our backbone network has three principal modules: two single-modal encoders (one for each modality), followed by a cross-modal encoder. All modules are based on a multi-layer Transformer~\cite{vaswani2017attention}. For the $\ell$-th Transformer layer, its output is
\begin{align} 
\hspace{-0mm}
\Hmat_{\ell} =  \Tcal (\Hmat_{\ell - 1}, \Hmat^{\prime}, \Mmat ) 
\label{eq_visual_emb}
\end{align}
where $\Hmat_{l-1} \in \R^{ L \times d_h}$ is the previous layer's features ($L$ is the sequence length), $\Hmat^{\prime} \in \R^{ L^{\prime} \times d_h}$ is the feature matrix to attend, and $\Mmat \in \R^{ L \times L^{\prime}} $ is the mask matrix, determining whether a pair of tokens can be attended to each other.
More specifically, in each Transformer block, the output vector is the concatenation of multiple attention heads $\Hmat_{\ell} = [\Amat_{\ell, 1}, \cdots, \Amat_{\ell, h}]$ ($h$ is the number of heads). One attention head $\Amat$ is computed via:
\begin{align} 
\hspace{-0mm}
& \Amat_{\ell}  = \text{Softmax}( \frac{  \Qmat \Kmat^{\top} }{ \sqrt{d_k} } + \Mmat )  \Vmat,\\
& \Mmat_{ij} = \left\{\begin{matrix}
0, & \text{allow to attend} \\ 
-\infty,  &  \text{not to attend}
\end{matrix}\right.  \\
& \Qmat = \Wmat^Q_{\ell} \Hmat^{\prime},  \Kmat = \Wmat^{K}_{\ell} \Hmat_{l-1},  \Vmat = \Wmat^V_{\ell} \Hmat_{l-1}
\label{eq_visual_emb}
\end{align}
where  $\Hmat_{l-1}$ and $\Hmat^{\prime}$ are linearly projected to a triple of queries, keys and values using parameter matrices $\Wmat^Q_{\ell}, \Wmat^{K}_{\ell}, \Wmat^V_{\ell} \in \R^{d_h \times d_k} $, respectively; $d_k$ is the projection dimension.
In the following, we use different mask matrices $\Mmat$ and attended feature matrices $\Hmat'$ to construct the contextualized representation for each module.

\paragraph{Single-modal Encoder} The standard self-attention layer is used in the single-modal encoder. All of the keys, values and queries come from the output of the previous layer in the encoder. Each position in the encoder can attend to all positions that belong to its own modality in the previous layer.
Specifically, $\Mmat$ is a full-zero matrix, and $\Hmat' = \Hmat_{l-1}$. Similar to the self-attention encoder module in the standard Transformer, the position-wise feed-forward network (FFN) is used.

\paragraph{Cross-modal Encoder} To fuse the features from both modalities, a cross-attention layer is considered. The queries $\Hmat'$ come from the previous layer of the other modality, and the memory keys and values come from the output $\Hmat_{l-1}$ of the current modality. It allows every position in the encoder to attend over all positions in the different modality. This mimics the typical encoder-decoder attention mechanisms in the Transformer, but here we consider two different modalities, rather than input-output sequences. This cross-attention layer is followed by a self-attention layer and an FFN layer. 

\begin{figure*}[t!]
	\vspace{-0mm}\centering
	\includegraphics[width=13.75cm]{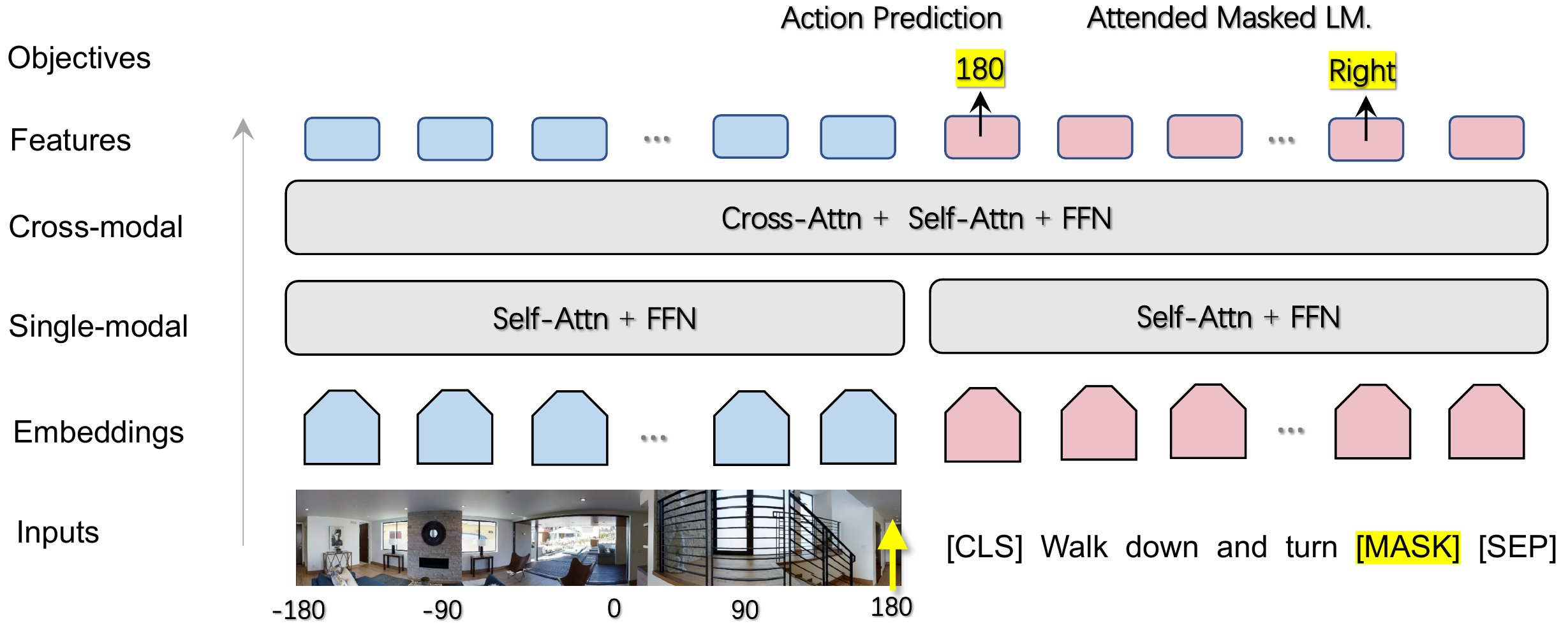}  
	\vspace{-0mm}
	\caption{Illustration of the proposed pre-training model. In this example, two learning objectives are considered: ($i$) image-attended masked language modeling is performed on the masked word $\mathtt{right}$ in the instruction; ($ii$) action prediction is performed to make the decision to navigate toward direction $\mathtt{180}$. Only the language features are used for fine-tuning in downstream tasks.}
	\vspace{-4mm}
	\label{fig:overall_scheme}
\end{figure*}

The overall model architecture is illustrated in Figure~\ref{fig:overall_scheme}. Following~\cite{tan2019lxmert}, $L_{\text{text}} = 9$,  $L_{\text{vision}} = 1$ and $L_{\text{cross}} = 3$. The last layer output of the encoder is denoted as $\zv = \hv_{L_{\text{cross}}}$, which is used as the features in the downstream tasks.

\subsection{Pre-training Objectives}

We introduce two main tasks to pre-train our model: Image-attended masked language modeling (MLM) and action prediction (AP). For an instruction-trajectory pair $\{\xv, \tauv\}$ from the training dataset $\Dcal_E$, we assume a state-action pair from the trajectory follows an independent identical distribution given the instruction in the pre-training stage: $ (\sv_t, \av_t) \overset{iid}{\sim} p(\tauv)$.

\paragraph{Attended Masked Language Modeling} 
We randomly mask out the input words with probability $15\%$, and replace the masked ones $x_i$ with special token
 $\mathtt{[MASK]}$. The goal is to predict these masked words based on the observation of their surrounding
words $\xv_{\backslash i}$ and all images $\sv$ by minimizing the negative log-likelihood:
\begin{align} 
\hspace{-0mm}
\Lcal_{\text{MLM}} = -\E_{  \sv \sim p(\tauv),  (\tauv, \xv) \sim \Dcal_E } \log p( x_i | \xv_{\backslash i}, \sv )
\label{eq_attend_mlm}
\end{align}
This is in analogy to the cloze task in BERT, where the masked word is recovered from surrounding words, but with additional image information to attend. It helps the learned word embeddings to be grounded in the context of visual states. This is particularly important for VLN tasks, where the agent is required to monitor the progress of completed instruction by understanding the visual images.

\paragraph{Action Prediction} 
The output on the special token $\mathtt{[CLS]}$ indicates the fused representation of both modalities. We apply an FC layer on top of the encoder output of $\mathtt{[CLS]}$ to predict the action. It scores how well the agent can make the correct decision conditioned on the current visual image and the instruction, without referring to the trajectory history. During training, we sample a state-action pair $(\sv, \av)$ from the trajectory $\tauv$ at each step, and then apply a cross-entropy loss for optimization:
\begin{align}
\Lcal_{\text{AP}} = -\E_{ (\av, \sv) \sim p(\tauv), (\tauv, \xv) \sim \Dcal_E } \log p( \av | x_{ \mathtt{[CLS]}}, \sv ).
\label{eq_action_prediction}
\end{align}
The full pre-training objective is: 
\begin{align}
\Lcal_{\text{Pre-training}} = \Lcal_{\text{MLM}}  + \Lcal_{\text{AP}}. 
\label{eq_pre_training}
\end{align}

\paragraph{Discussion} Other loss designs can be considered for the pre-training objective. Our results on masked image modeling did not show better results, and thus are excluded in the experiments.

\subsection{Pre-training Datasets}
We construct our pre-training dataset based on the Matterport3D Simulator, a photo-realistic visual reinforcement learning (RL) simulation environment for the development of intelligent agents based on the Matterport3D dataset~\cite{chang2017matterport3d}.
Specifically, it consists of two sets: 
$(\RN{1})$ The training datasets of R2R, which has 104K image-text-action triplets; 
$(\RN{2})$ we employed the Speaker model in~\cite{fried2018speaker} to synthesize 1,020K instructions for the shortest-path trajectories on the training environments. This leads to 6,482K image-text-action triplets. Therefore, the pre-training dataset size is 6,582K.

\section{Adapting to new tasks}
We focus on three downstream VLN tasks that are based on the Matterport3D simulator.
Each task poses a very different challenge to evaluate the agent. 
$(\RN{1})$ The R2R task is used as an in-domain task; it can verify the agent's generalization capability to unseen environments. 
$(\RN{2})$ CVDN and HANNA are considered as out-of-domain tasks, to study the generalization ability of the agent to new tasks. More specifically, CVDN considers indirect instructions (\ie, dialog history), and HANNA is an interactive RL task.

\subsection{Room-to-Room}
In R2R, the goal is to navigate from a starting position to a target position with the minimal trajectory length, where the target is explicitly informed via language instruction. To use the pre-trained model for fine-tuning in R2R, the attended contextualized word embeddings are fed into an LSTM encoder-decoder framework, as in~\cite{fried2018speaker,li2019robust}. In prior work, random initialization is used in~\cite{fried2018speaker}, and BERT is used in~\cite{li2019robust}. In contrast, our word embeddings are pre-trained from scratch with VLN data and tasks. 

\subsection{Cooperative Vision-and-Dialogue Navigation}
In the CVDN environment, the Navigation from Dialog History (NDH) is defined, where the agent searches an environment for a goal location, based on the dialog history that consists of multiple turns of question-answering interactions between the the agent and to its partner. The partner has privileged access to the best next steps that the agent should take according to a shortest path planner. CVDN is more challenging than R2R, in that the instructions from the dialog history are often ambiguous, under-specified, and indirect to the final target. The fine-tuning model architecture for CVDN is the same as R2R, except that CVND usually has much longer text input. We limit the sequence length to 300. Words that are longer than 300 in a dialog history are removed.

\subsection{HANNA: Interactive Imitation Learning} 
HANNA simulates a scenario where a human requester asks an agent via language to find an object in an indoor environment, without specifying the process of how to complete the task. The only source of help the agent can leverage in the environment is the {\em assistant}, who helps the agent by giving subtasks in the form of 
$(\RN{1})$ a natural language instruction that guides the agent to a specific location, and 
$(\RN{2})$ an image of the view at that location. When the help mode is triggered, we use our pre-trained model to encode the language instructions, and the features are used for the rest of their system.

\section{Experimental Results}

\subsection{Training details}
\paragraph{Pre-training} 
We pre-train the proposed model on eight V100 GPUs, and the batch size for each GPU is 96. The AdamW optimizer~\cite{kingma2014adam} is used, and the learning rate is $5\!\times\!10^{-5}$. The total number of training epochs is 20.

\paragraph{Fine-tuning} The fine-tuning is performed on NVIDIA 1080Ti GPU.
For the R2R task, we follow the same learning schedule as~\cite{tan2019learning}. When training the augmented listener, we use batch size 20. We continue to fine-tune the cross-attention encoder for 20k iterations, with the batch size 10 and learning rate $2\!\times\!10^{-6}$. 
For the NDH task, we follow the same learning schedule as in~\cite{thomason2019vision}, and choose the batch size as 15 and learning rate as $5\!\times\!10^{-4}$. 
For HANNA, the training schedule is the same as~\cite{nguyen2019help}. The batch size is 32 and learning rate is $1\!\times\!10^{-4}$.

\subsection{Room-to-Room}
\paragraph{Dataset} The R2R dataset~\cite{anderson2018vision} consists of 10,800 panoramic views (each panoromic view has 36 images) and 7,189 trajectories. Each trajectory is paired with three natural language instructions. The R2R dataset consists of four splits: train, validation seen and validation unseen, test unseen. The challenge of R2R is to test the agent's generalization ability in unseen environments. 

\paragraph{Evaluation Metrics}
The performance of different agents is evaluated using the following metrics:
\begin{itemize}[noitemsep,topsep=2pt] 
\item[\textbf{\texttt{TL}}] \textbf{Trajectory Length} measures the average length of the navigation trajectory.
\item[\textbf{\texttt{NE}}] \textbf{Navigation Error} is the mean of the shortest path distance in meters between the agent's final location and the target location. 
\item[\textbf{\texttt{SR}}] \textbf{Success Rate} is the percentage of the agent's final location that is less than 3 meters away from the target location.
\item[\textbf{\texttt{SPL}}] \textbf{Success weighted by Path Length~\cite{anderson2018evaluation}} trades-off \texttt{SR} against \texttt{TL}. A higher score represents more efficiency in navigation.
\end{itemize}
Among these metrics, \texttt{SPL} is the recommended primary metric, and other metrics are considered as auxiliary measures.

\begin{table*}[ht!]
\small
\centering
\begin{tabular}{@{\hspace{3pt}}l@{\hspace{3pt}}l@{}r@{\hspace{9pt}}c@{\hspace{9pt}}c@{\hspace{9pt}}c|r@{\hspace{9pt}}c@{\hspace{9pt}}c@{\hspace{9pt}}c|r@{\hspace{9pt}}c@{\hspace{9pt}}c@{\hspace{9pt}}c}
\toprule
& & \multicolumn{4}{c}{Validation Seen} & \multicolumn{4}{c}{Validation Unseen} & \multicolumn{4}{c}{Test Unseen} \\ 
& Agent & \texttt{TL} $\downarrow$ & \texttt{NE} $\downarrow$ & \texttt{SR} $\uparrow$ & \texttt{SPL} $\uparrow$ & \texttt{TL} $\downarrow$  & \texttt{NE} $\downarrow$ & \texttt{SR} $\uparrow$ & \texttt{SPL} $\uparrow$ & \texttt{TL} $\downarrow$ & \texttt{NE} $\downarrow$ & \texttt{SR} $\uparrow$ & \texttt{SPL} $\uparrow$\\ 
\midrule
& \textsc{Random} & 9.58 & 9.45 & 16 & - & 9.77 & 9.23 & 16 & - & 9.93 & 9.77 & 13 & 12 \\
& \textsc{Seq2Seq} & 11.33 & 6.01 & 39 & - & 8.39 & 7.81 & 22 & - & \phantom{0,0}8.13 & 7.85 & 20 & 18 \\
& \textsc{RPA} & - & 5.56 & 43 & - & - & 7.65 & 25 & - & \phantom{0,0}9.15& 7.53 & 25 & 23 \\
\multirow{2}{*}{\rotatebox{90}{\footnotesize Greedy, {\bf S}}} & \textsc{Speaker-Follower} & - & 3.36 & 66 & - & - & 6.62 & 35 & - & \phantom{0,0}14.82 & 6.62 & 35 & 28\\
& \textsc{SMNA} & - & - & - & - & - & - & - & - & \phantom{0,0}18.04 & 5.67 & 48 & 35 \\
& \textsc{RCM+SIL(train)} & 10.65 & 3.53 & 67 & - & 11.46  & 6.09 & 43 & - & 11.97 & 6.12 & 43 & 38 \\
& \textsc{Regretful} & - & 3.23 & 69 & 63 & - & 5.32 & 50 & 41 & 13.69 & 5.69 & 48 & 40 \\
& \textsc{Fast} & - & - & - & - & 21.17 & 4.97 & 56 & 43 & 22.08 & 5.14 & 54 & 41 \\
& \textsc{EnvDrop} & 11.00 & 3.99 & 62 & 59 & 10.70 & 5.22 & 52 & 48 & 11.66 & 5.23 & 51 & 47 \\

\cellcolor{white} & \textsc{Press}  & 10.57 & 4.39 & 58 & 55 & 10.36 & 5.28 & 49 & 45 & 10.77 & 5.49 & 49 & 45 \\

%
\rowcolor{Gray}
\cellcolor{white} & \textsc{Prevalent} (ours)  & 10.32 & 3.67 & 69 & \textcolor{blue}{\textbf{65}} & 10.19 & \textcolor{blue}{\textbf{4.71}} & \textcolor{blue}{\textbf{58}} & \textcolor{blue}{\textbf{53}} & 10.51 & 5.30 & \textcolor{blue}{\textbf{54}} & \textcolor{blue}{\textbf{51}} \\
\midrule
\multirow{2}{*}{\footnotesize {\bf M}} & 
\textsc{Press}  & 10.35 &  3.09 & 71 & 67 & 10.06 & 4.31 &  59 & 55 & 10.52 & 4.53 & 57 & 53 \\
& \textsc{Prevalent}  & \textcolor{blue}{\textbf{10.31}}  & 3.31 & 67 & 63 & \textcolor{blue}{\textbf{9.98}}  & \textcolor{blue}{\textbf{4.12}} & \textcolor{blue}{\textbf{60}} & \textcolor{blue}{\textbf{57}} & \textcolor{blue}{\textbf{10.21}}  & \textcolor{blue}{\textbf{4.52}} & \textcolor{blue}{\textbf{59}} & \textcolor{blue}{\textbf{56}} \\
%
%
\midrule
%
& Human & - & - & - & - & - & - & - & - & \phantom{0,0}11.85 & 1.61 & 86 & 76 \\
\bottomrule
\end{tabular}
\vspace{-1mm}
\caption{
Comparison with the state-of-the-art methods on R2R. \textcolor{blue}{Blue} indicates the best value in a given setting. {\textbf{S}} indicates the single-instruction setting, {\textbf{M}} indicates the multiple-instruction setting.
}
\label{tab:main_result_r2r}
\vspace{-4mm}
\end{table*}

\paragraph{Baselines}
We compare our approach with \emph{nine} recently published systems:
\begin{itemize}[noitemsep,topsep=0pt]
\item \textsc{Random}: an agent that randomly selects a direction and moves five step in that direction ~\cite{anderson2018vision}. 
\item \textsc{S2S-Anderson}: a sequence-to-sequence model using a limited discrete action space~\cite{anderson2018vision}.
\item \textsc{RPA}~\cite{wang2018look}: an agent that combines model-free and model-based reinforcement learning, using a look-ahead module for planning. 
\item \textsc{Speaker-Follower}~\cite{fried2018speaker}: an agent trained with data augmentation from a speaker model on the panoramic action space.
\item \textsc{Smna}~\cite{ma2019self}: an agent trained with a visual-textual co-grounding module and a progress monitor on the panoramic action space.
\item \textsc{RCM+SIL}~\cite{wang2018reinforced}: an agent trained with cross-modal grounding locally and globally via RL.
\item \textsc{Regretful}~\cite{ma2019regretful}: an agent with a trained progress monitor heuristic for search that enables backtracking. 
\item \textsc{Fast}~\cite{ke2019tactical}: an agent that uses a fusion function to score and compare partial trajectories of different lengths, which enables the agent to efficiently backtrack after a mistake.
\item \textsc{EnvDrop}~\cite{tan2019learning}: an agent is trained with environment dropout, which can generate more environments based on the limited seen environments. 
\item \textsc{PreSS}~\cite{li2019robust}: an agent is trained with pre-trained language models and stochastic sampling to generalize well in the unseen environment.
\end{itemize}

\paragraph{Comparison with SoTA}
Table~\ref{tab:main_result_r2r} compares the performance of our agent against the existing published top systems.\footnote{The full list of leaderboard is publicly available: \url{https://evalai.cloudcv.org/web/challenges/challenge-page/97/leaderboard/270}}. Our agent \short{} outperforms the existing models on SR and SPL by a large margin. On both validation seen and unseen environments, \short{} outperforms other agents on nearly all metrics. 

In \textsc{PreSS}~\cite{li2019robust}, multiple introductions are used. To have a fair comparison, we follow~\cite{li2019robust}, and report \short{} results. We see that testing SPL is improved. Further, the gap between seen and unseen environments of \short{} is smaller than \textsc{PreSS}, meaning that image-attended language understanding is more effective to help the agent generalize better to an unseen environment.

\begin{table*}[ht!]
\small
\centering
\begin{tabular}{@{\hspace{3pt}}l@{\hspace{3pt}}l@{}r@{\hspace{9pt}}c@{\hspace{9pt}}c|r@{\hspace{9pt}}c@{\hspace{9pt}}c r@{\hspace{9pt}}c@{\hspace{9pt}}c }
\toprule
&  & \multicolumn{3}{c}{Validation Unseen} & \multicolumn{3}{c}{Test Unseen} \\ 
& Agent  & \textbf{Oracle}  & \textbf{Navigator} & \textbf{Mixed}  & \textbf{Oracle} & \textbf{Navigator} & \textbf{Mixed}\\ 
\midrule
& \textsc{Random}  & 1.09 & 1.09 & 1.09 & 0.83 & 0.83 & 0.83  \\
& \textsc{Seq2Seq}  & 1.23 & 1.98 & 2.10 & 1.25 & 2.11 & 2.35  \\
\rowcolor{Gray}
\cellcolor{white}
&  \short{} (Ours)  &  \textcolor{blue}{\textbf{2.58}}  & \textcolor{blue}{\textbf{2.99}}  &  \textcolor{blue}{\textbf{3.15}} 
& \textcolor{blue}{\textbf{1.67}}  & \textcolor{blue}{\textbf{2.39}} & \textcolor{blue}{\textbf{2.44}} \\

\midrule
& \textsc{Shortest Path Agent} & 8.36 & 7.99 & 9.58 & 8.06 & 8.48 & 9.76 \\

%

%
\bottomrule
\end{tabular}
\vspace{-1mm}
\caption{Results on CVDN measured by Goal Progress. \textcolor{blue}{Blue} indicates the best value in a given setting.
}
\label{tab:main_result_cvdn}
\vspace{-0mm}
\end{table*}

\subsection{Cooperative Vision-and-Dialogue Navigation}
\paragraph{Dataset \& Evaluation Metric} 
The CVDN dataset has 2050 human-human navigation dialogs, comprising over 7K navigation trajectories punctuated by question-answer exchanges, across 83 MatterPort houses~\cite{chang2017matterport3d} .
The metrics for R2R can be readily used for the CVDN dataset. Further, one new metric is proposed for the NDH task: 
\begin{itemize}[noitemsep,topsep=2pt] 
\item[\textbf{\texttt{GP}}] \textbf{Goal Progress} measures the difference between completed distance and left distance to the goal. Larger values indicate a more efficient agent.
\end{itemize}

Three settings are considered, depending on which ground-truth action/path is employed~\cite{thomason2019vision}. {\it Oracle} indicates the shortest path, and {\it Navigator} indicates the path taken by the navigator. The {\it Mixed} supervision path means to take the navigator path if available, otherwise the shortest path.
The results are in Table~\ref{tab:main_result_cvdn}. The proposed \short{} significantly outperforms the Seq2Seq baseline on both validation and testing unseen environments in all settings, leading to the top position on the leaderboard~\footnote{The full list of leaderboard is publicly available: \url{https://evalai.cloudcv.org/web/challenges/challenge-page/463/leaderboard/1292}}. 
Note that our encoder is pre-trained on R2R dataset. We observe that it can provide significant improvement when used the new task built on the CVDN dataset. This shows that the pre-trained model can adapt well on new tasks, and yields better generalization.

\begin{table*}[ht!]
\small
\centering
\begin{tabular}{@{\hspace{3pt}}l@{\hspace{3pt}}lr@{\hspace{9pt}}c@{\hspace{9pt}}c@{\hspace{9pt}}c|r@{\hspace{9pt}}c@{\hspace{9pt}}c@{\hspace{9pt}}c}
\toprule
& & \multicolumn{4}{c}{SEEN-ENV} & \multicolumn{4}{c}{UNSEEN-ALL} \\ 
& Agent  & \texttt{SR} $\uparrow$  & \texttt{SPL} $\uparrow$ & \texttt{NE} $\downarrow$ & \texttt{\#R} $\downarrow$ & \texttt{SR} $\uparrow$ & \texttt{SPL} $\uparrow$ & \texttt{NE} $\downarrow$ & \texttt{\#R} $\downarrow$\\ 
\midrule
\multirow{2}{*}{ \rotatebox{90}{\footnotesize {Rule}} }
& \textsc{Random Walk}  & 0.54 & 0.33 & 15.38 & 0.0 & 0.46 & 0.23 & 15.34 & 0.0 \\
& \textsc{Forward 10}  & 5.98 & 4.19 & 14.61 & 0.0 & 6.36 & 4.78 & 13.81 & 0.0 \\
\midrule
\multirow{2}{*}{}
& \textsc{No assistance}  & 17.21 & 13.76 & 11.48 & 0.0 & 8.10 & 4.23 & 13.22 & 0.0 \\
& \textsc{Anna}  & 88.37 & 63.92 & 1.33 & 2.9 & 47.45 & 25.50 & 7.67 & 5.8 \\ 
\rowcolor{Gray}
\cellcolor{white}
& \short{} (Ours)  & 83.82 & 59.38 & 1.47 & 3.4 & \textcolor{blue}{\textbf{52.91}} & \textcolor{blue}{\textbf{28.72}} & \textcolor{blue}{\textbf{5.29}} & 6.6 \\
\midrule
\multirow{2}{*}{ \rotatebox{90}{\footnotesize { Skyline}} }
& \textsc{Shortest} & 100.00& 100.00 & 0.00 & 0.0 & \phantom{0,0}100.00 & 100.00 & 0.00 & 0.0 \\
& Perfect assistance & 90.99 & 68.87 & 0.91 & 2.5 & \phantom{0,0}83.56 & 56.88 & 1.83 & 3.2 \\

%
\bottomrule
\end{tabular}
\vspace{-1mm}
\caption{
Results on test splits of HANNA. The agent with ``perfect assistance'' uses the teacher navigation policy to make decisions when executing a subtask from the assistant. \textcolor{blue}{Blue} indicates the best value.
}
\label{tab:main_result_hanna}
\vspace{-2mm}
\end{table*}

\subsection{HANNA}

\paragraph{Dataset \& Evaluation Metric} The HANNA dataset features 289 object types; the language instruction vocabulary contains 2,332 words. The numbers of locations on the shortest paths to the requested objects are restricted to be between 5 and 15. With an average edge length of 2.25 meters, the agent has to travel about 9 to 32 meters to reach its goals. Similar to R2R, SR, SPL and NE are used to evaluate the navigation. Further, one new metric is considered for this interactive task:
\begin{itemize}[noitemsep,topsep=2pt] 
\item[\textbf{\texttt{\#R}}] \textbf{Number of requests} measures how many helps are requested by the agent. 
\end{itemize}

The results are shown in Table~\ref{tab:main_result_hanna}. Two rule-based methods and two skyline methods are reported as references; see~\cite{nguyen2019help} for details.
Our \short{} outperforms the baseline agent \textsc{Anna} on the test unseen environments in terms of SR, SPL and NE, while requesting a slightly higher number of helps (\#R). When measuring the performance gap between seen and unseen environments, we see that \short{} shows a significantly smaller difference than \textsc{Anna}, \eg, (59.38-28.72=30.66) vs (63.92-25.50=38.42) for SPL. This means that the pre-trained joint representation by \short{} can reduce over-fitting, and generalise better to unseen environments.

\subsection{Ablation Studies}

\begin{table*}[ht!]
\small
\centering
\begin{tabular}{ll@{}r@{\hspace{9pt}}c@{\hspace{9pt}}c|r@{\hspace{9pt}}c@{\hspace{9pt}}c|r@{\hspace{9pt}}c@{\hspace{9pt}}c r@{\hspace{9pt}}c@{\hspace{9pt}}c }
\toprule
&  & \multicolumn{3}{c}{Navigation QA
} & \multicolumn{3}{c}{Oracle Answer}   & \multicolumn{3}{c}{All}\\ 
& Methods  & \textbf{Oracle}  & \textbf{Navigator} & \textbf{Mixed}  & \textbf{Oracle} & \textbf{Navigator} & \textbf{Mixed}  & \textbf{Oracle} & \textbf{Navigator} & \textbf{Mixed}\\ 
\midrule
& $\Lcal_{\text{PA}} + \Lcal_{\text{MLM}} $  & \textcolor{blue}{\textbf{2.80}}  & \textcolor{blue}{\textbf{3.01}}   &	\textcolor{blue}{\textbf{3.28}}   & 2.78	 & \textcolor{blue}{\textbf{3.44}}  & \textcolor{blue}{\textbf{3.38}}  & \textcolor{blue}{\textbf{2.58}}   &	\textcolor{blue}{\textbf{2.99}}   &	\textcolor{blue}{\textbf{3.15}}    \\
& $\Lcal_{\text{MLM}} $  & 2.69 &	3.00 &	3.25 & 2.84 &	3.35 &	3.19 & 2.52 &	2.98 &	3.14 \\
&  BERT pre-trainig  & 2.26 &	2.71 &	2.94 & 2.70 &	2.68 &	3.06 & 2.46 &	2.74 &	2.64 \\
&  BERT fine-tuning  & 2.39 &	2.03 &	2.51 & 2.23	& 2.41 &	2.52 & 2.32	& 2.93 &	2.28
\\
\bottomrule
\end{tabular}
\vspace{-1mm}
\caption{Ablation study of the pre-training objectives on CVDN, measured by Goal Progress. \textcolor{blue}{Blue} indicates the best value.
}
\label{tab:result_cvdn_ablation_action}
\vspace{-2mm}
\end{table*}

\begin{table*}[ht!]
\small
\centering
\begin{tabular}{@{\hspace{5pt}}l@{\hspace{5pt}}l@{}r@{\hspace{12pt}}c@{\hspace{9pt}}c@{\hspace{9pt}}c|r@{\hspace{9pt}}c@{\hspace{9pt}}c@{\hspace{9pt}}c|r@{\hspace{9pt}}c@{\hspace{9pt}}c@{\hspace{9pt}}c}
\toprule
& & \multicolumn{4}{c}{Validation Seen} & \multicolumn{4}{c}{Validation Unseen} & \multicolumn{4}{c}{Test Unseen} \\ 
& Methods & \texttt{TL} $\downarrow$ & \texttt{NE} $\downarrow$ & \texttt{SR} $\uparrow$ & \texttt{SPL} $\uparrow$ & \texttt{TL} $\downarrow$  & \texttt{NE} $\downarrow$ & \texttt{SR} $\uparrow$ & \texttt{SPL} $\uparrow$ & \texttt{TL} $\downarrow$ & \texttt{NE} $\downarrow$ & \texttt{SR} $\uparrow$ & \texttt{SPL} $\uparrow$\\ 
\midrule
& Two-stage & ~~~10.32	&  \textcolor{blue}{\textbf{ 3.67}}	& \textcolor{blue}{\textbf{ 0.69 }} & 	   \textcolor{blue}{\textbf{0.66}}& 	
10.19  & \textcolor{blue}{\textbf{4.71}} 	& \textcolor{blue}{\textbf{0.58}}  	& 	\textcolor{blue}{\textbf{0.53}}  	& 	  
10.51	
&  \textcolor{blue}{\textbf{5.30}} 	& 	\textcolor{blue}{\textbf{0.54}}  	& 	\textcolor{blue}{\textbf{0.51}}   \\
& Feature-based & ~~~10.13	& 3.98	& 0.66	& 0.64	& 9.70	& 5.01	& 0.54	& 0.51	& 9.99	& 5.54	& 0.52	& 0.49 \\
\bottomrule
\end{tabular}
\vspace{-1mm}
\caption{Ablation study on R2R: feature-based vs fine-tuning. \textcolor{blue}{Blue} indicates the better value.}
\label{tab:ablation_result_finetunig_r2r}
\vspace{-2mm}
\end{table*}

\paragraph{Is pre-training with actions helpful?} 
Our pre-training objective in~\eqref{eq_pre_training} includes two losses, $\Lcal_{\text{PA}} $ and $ \Lcal_{\text{MLM}} $. To study the impact of each loss, we pre-train two model variants: one is based on the full objective $ \Lcal_{\text{PA}} + \Lcal_{\text{MLM}} $, the other only uses $ \Lcal_{\text{MLM}} $. To verify its impact on new tasks, we consider CVDN first, and the results are shown in Table~\ref{tab:result_cvdn_ablation_action}. Three types of text inputs are considered: Navigation QA, Orcale Answer, and All (a combination of both). More details are provided in the Appendix.

When $\Lcal_{\text{PA}}$ is employed in the objective, we see consistent improvement on nearly all metrics and settings. 
Note that our MLM is different from BERT in that the attention over images is used in the cross-layer. To verify whether the image-attended learning is necessary, we consider BERT in two ways. 
$(\RN{1})$ BERT pre-training: we apply the original MLM loss in BERT on our R2R pre-training dataset. The newly pre-trained BERT is used for fine-tuning on CVDN. 
$(\RN{2})$ BERT fine-tuning: we directly fine-tune off-the-shelf BERT on CVDN.  
Their performances are lower than the two variants of the proposed \short{}. This means our image-attended MLM is more effective for navigation tasks. More ablation studies on the pre-training objectives are conducted for HANNA, with results shown in the Appendix.

\paragraph{Feature-based vs Fine-tuning}

The pre-trained encoder can be used in two modes: 
$(\RN{1})$ {\em fine-tuning} approach, where a task-specific layer is added to the pre-trained model, and all parameters are jointly updated on a downstream task. 
$(\RN{2})$ {\em feature-based} approach, where fixed features are extracted from the pre-trained model, and only the task-specific layer is updated.
In this paper, all \short{} presented results generally have used the feature-based approach, as there are major computational benefits to pre-computing an expensive representation of the training data once, and then running many experiments with cheaper models on top of this representation. In the R2R dataset, we consider a {\em two-stage} scheme, where we fine-tune the cross-attention layers of the agent, after training via the feature-based approach. The results are reported in Table~\ref{tab:ablation_result_finetunig_r2r}. We observe notable improvement with this two-stage scheme on nearly all metrics, expect the trajectory length.

\begin{figure}[t!]
	\vspace{-0mm}\centering
	\begin{tabular}{c}
		\hspace{-4mm}
		\includegraphics[height=3.6cm]{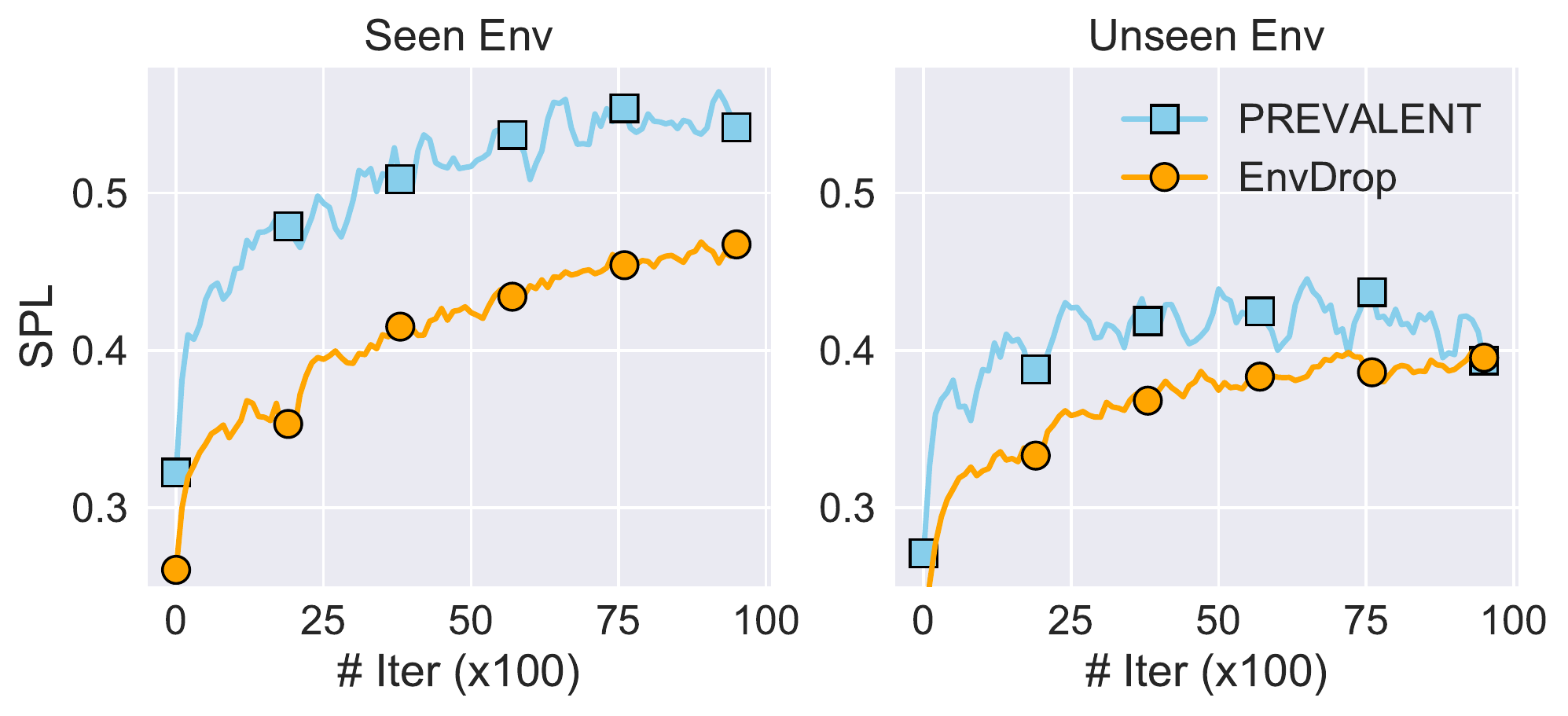} 
		\vspace{-2mm}\\
		(a) R2R  \\
		\hspace{-2mm}
		\includegraphics[height=3.6cm]{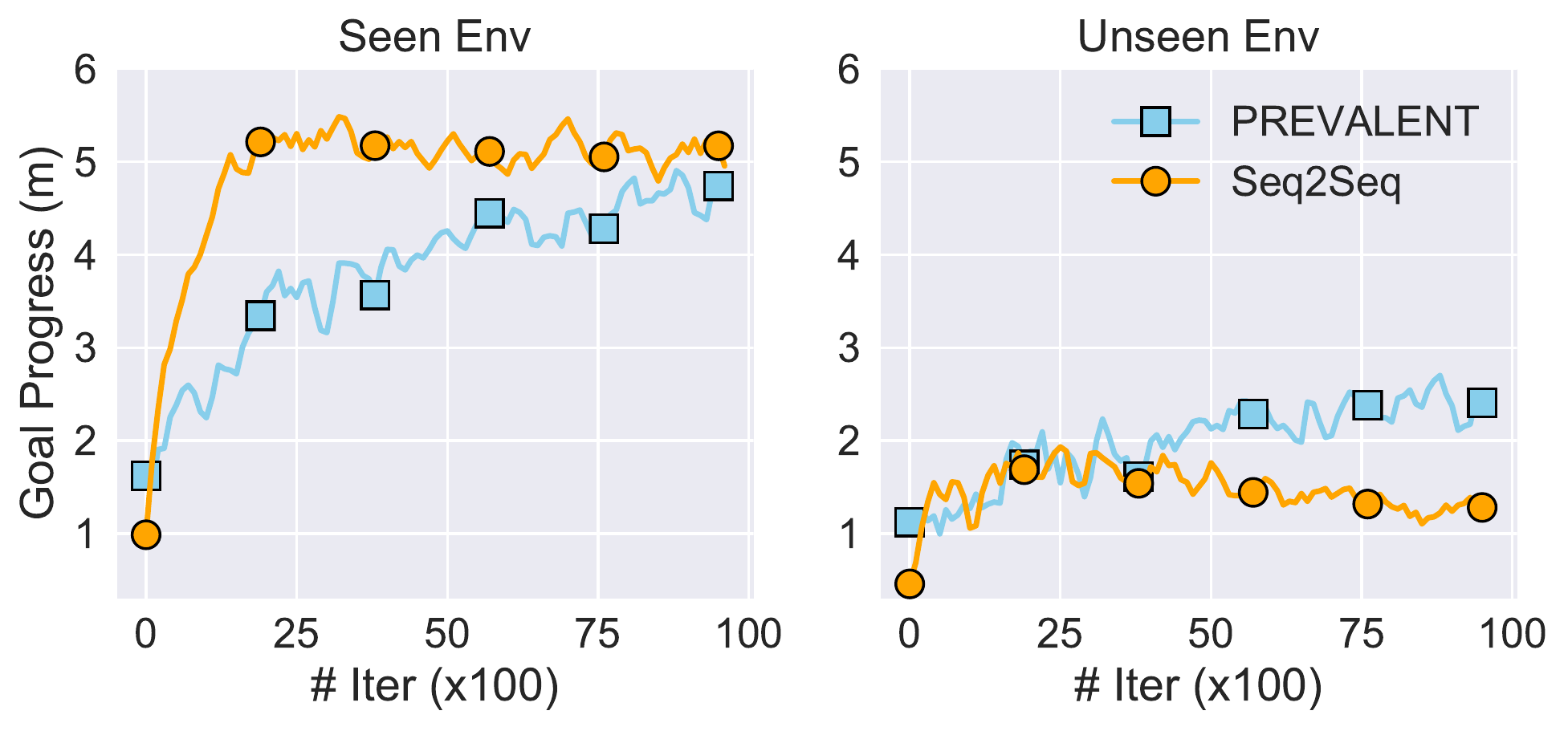} 
		\vspace{-2mm} \\
		(b) CVDN 
		 \vspace{-2mm}
	\end{tabular}
	\caption{Learning curves on (a) R2R and (b) CVDN.}
	\vspace{-0mm}
	\label{fig:lc_comparison}
\end{figure}

\paragraph{How does pre-training help generalization?} We plot the learning curves on the seen/unseen environments for R2R in Figure~\ref{fig:lc_comparison}(a), and CVDN in Figure~\ref{fig:lc_comparison}(b). 
Compared with the random initialized word embeddings in EnvDrop~\cite{tan2019learning}, the pre-trained word embeddings can adapt faster (especially in the early stage), and converge to higher performance in unseen environments. This is demonstrated by the SPL values in the Figure~\ref{fig:lc_comparison}(a).
By comparing the learning curves in Figure~\ref{fig:lc_comparison}(b), we see a much smaller gap between seen and unseen environments for \short{} than the Seq2Seq baseline~\cite{thomason2019vision}, meaning pre-training is an effective tool to help reduce over-fitting in learning.

\section{Conclusions}
We present \short{}, a new pre-training and fine-tuning paradigm for vision-and-language navigation problems. This allows for more effective use of limited training data to improve generalization to previously unseen environments, and new tasks. The pre-trained encoder can be easily plugged into existing models to boost their performance. Empirical results on three benchmarks (R2R, CVDN and HANNA) demonstrate that \short{} significantly improves over existing methods, achieving new state-of-the-art performance.

\newpage

{\small
\bibliographystyle{ieee_fullname}
\bibliography{egbib}
}

\onecolumn
\appendix
\newpage

\begin{center}
	\bf{\Large Supplementary Material: Towards Learning a Generic Agent for \\ Vision-and-Language Navigation via Pre-training} \\ 
	\vspace{5mm}
\end{center}

\paragraph{Summary of Contributions.} Weituo implemented the algorithm, made the model work, and ran all experiments. Chunyuan initiated the idea of pre-training the first generic agent for VLN, led and completed the manuscript writing. Xiujun provided the codebase and helped implementation. Lawrence and Jianfeng edited the final manuscript.

\begin{wrapfigure}{R}{8.8cm}
\vspace{-4mm}
\centering
{
\includegraphics[width=3.5in]{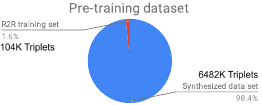}
}
\caption{The percentage of pre-training datasets. The synthesized dataset occupies 98.4\%.}
\label{exp:nm}
\vspace{-5mm}
\end{wrapfigure}
\section{Pre-training Dataset Preparation}
We found that the largest VLN training dataset R2R contains only 104K samples, an order magnitude smaller than the pre-training datasets typically used in language~\cite{devlin2018bert} or vision-and-language pre-training~\cite{zhou2019unified}. This renders a case where pre-training can be degraded due to insufficient training data, while harvesting such samples with human annotations is expensive. Fortunately, we can resort to generative models to synthesize the samples. We first train an seq2seq auto-regressive model (\ie, a speaker model~\cite{fried2018speaker}) that can produce language instructions conditioned on the agent trajectory (a sequence of actions and visual images) on R2R dataset; then collect a large number of shortest trajectories using the Matterport 3D Simulator, and synthesize their corresponding instructions using the speaker model. This leads to 6482K new training samples. The two datasets are compared in Figure 4(b). The agent is pre-trained on the combined dataset. Our results show that synthetic samples produced by generative models can be incorporated into the pre-training data and helps self-supervised learning.

\section{Experiments}

\paragraph{Three types of inputs on CVDN} We illustrate the  naming of three types of text inputs on CVDN in Table~\ref{tab:naming_cvdn}.

\begin{table*}[ht!]
\small
\centering
\begin{tabular}{@{\hspace{3pt}}l@{\hspace{3pt}}lr@{\hspace{9pt}}c@{\hspace{9pt}}c@{\hspace{9pt}}c@{\hspace{9pt}}c}
\toprule
&  &  $V$  & $t_0$ &  $A_i$ & $Q_i$ & $Q_{1:i-1} \& A_{1:i-1}$ \\ 
\midrule
& Oracle Answer &  $\checkmark$  &  $\checkmark$  & $\checkmark$ & &\\
& Navigation QA   & $\checkmark$ &  $\checkmark$  &  $\checkmark$  & $\checkmark$ & \\
& All  &  $\checkmark$  &  $\checkmark$  &  $\checkmark$  &  $\checkmark$   &  $\checkmark$ \\
\bottomrule
\end{tabular}
\vspace{-1mm}
\caption{
Three types of inputs on CVDN. $t_0$ is the target object, $V$ is the ResNet feature. $Q_i$ and $A_i$ are the question and answers in the $i$-th turn.  $Q_{1:i-1} \& A_{1:i-1}$ are the question \& answer pairs before the $i$-th turn. 
}
\label{tab:naming_cvdn}
\vspace{-2mm}
\end{table*}

\paragraph{Ablation Study Results on HANNA} Table~\ref{tab:ablation_result_hanna} shows the results with different pre-training objectives. We see that the $ \Lcal_{\text{PA}} + \Lcal_{\text{MLM}} $ yields the best performance among all variants.

\begin{table*}[ht!]
\small
\centering
\begin{tabular}{@{\hspace{3pt}}l@{\hspace{3pt}}lr@{\hspace{9pt}}c@{\hspace{9pt}}c@{\hspace{9pt}}c|r@{\hspace{9pt}}c@{\hspace{9pt}}c@{\hspace{9pt}}c}
\toprule
& & \multicolumn{4}{c}{SEEN-ENV} & \multicolumn{4}{c}{UNSEEN-ALL} \\ 
& Agent  & \texttt{SR} $\uparrow$  & \texttt{SPL} $\uparrow$ & \texttt{NE} $\downarrow$ & \texttt{\#R} $\downarrow$ & \texttt{SR} $\uparrow$ & \texttt{SPL} $\uparrow$ & \texttt{NE} $\downarrow$ & \texttt{\#R} $\downarrow$\\ 
\midrule
\rowcolor{Gray}
\cellcolor{white}
& \short{} ($\Lcal_{\text{PA}} + \Lcal_{\text{MLM}} $)  & 
\textcolor{blue}{\textbf{83.82}}  & \textcolor{blue}{\textbf{59.38}}  & 
\textcolor{blue}{\textbf{1.47}}  & \textcolor{blue}{\textbf{3.4}}  & \textcolor{blue}{\textbf{52.91}} & \textcolor{blue}{\textbf{28.72}} & \textcolor{blue}{\textbf{5.29}} & \textcolor{blue}{\textbf{6.6}}  \\
& \short{} ($\Lcal_{\text{MLM}} $) & 78.75 & 54.68 & 1.82 & 4.3 & 44.29 & 24.27 & 6.33 & 8.1 \\
& \textsc{BERT} (feature-based) & 57.54 & 34.33 & 4.71 & 3.9 & 24.12 & 11.50 & 9.55 & 11.3 \\
& \textsc{BERT} (fine-tuning)  & 80.75 & 57.46 & 1.97 & 4.0 & 26.36 & 12.66 & 9.1 & 8.3 \\
\bottomrule
\end{tabular}
\vspace{-1mm}
\caption{
Ablation study of pre-training objectives on test splits of HANNA.
}
\label{tab:ablation_result_hanna}
\vspace{-2mm}
\end{table*}

\section{Comparison with Related Work}
\paragraph{Comparison with \textsc{Press}.} 
The differences are summarized in Table~\ref{tab:comparison} (a). Empirically, we show that (1) incorporating visual and action information into pre-training can improve navigation performance; (2) Pre-training can generalize across different new navigation tasks.
\vspace{-0mm}


\vspace{-0mm}
\paragraph{Comparison with vision-language pre-training (VLP).} 
The differences are in Table~\ref{tab:comparison} (b). Though the proposed methodology generally follows self supervised learning such as VLP or BERT, our research scope and problem setups are different, which renders existing pre-models are not readily applicable.

\begin{table}[h!]\centering \hspace{-2.5mm}
	\begin{minipage}{1.0\linewidth}
		\vskip 0.0in
		\centering
		\begin{tabular}{cc}
        \includegraphics[width=8.45cm]{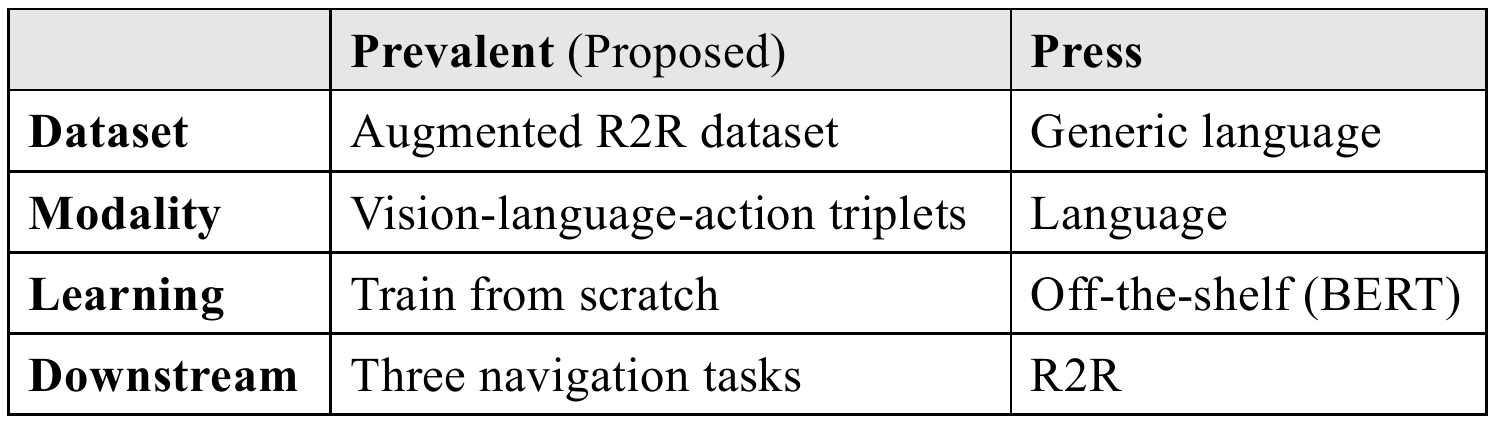}
        & 
        \includegraphics[width=8.45cm]{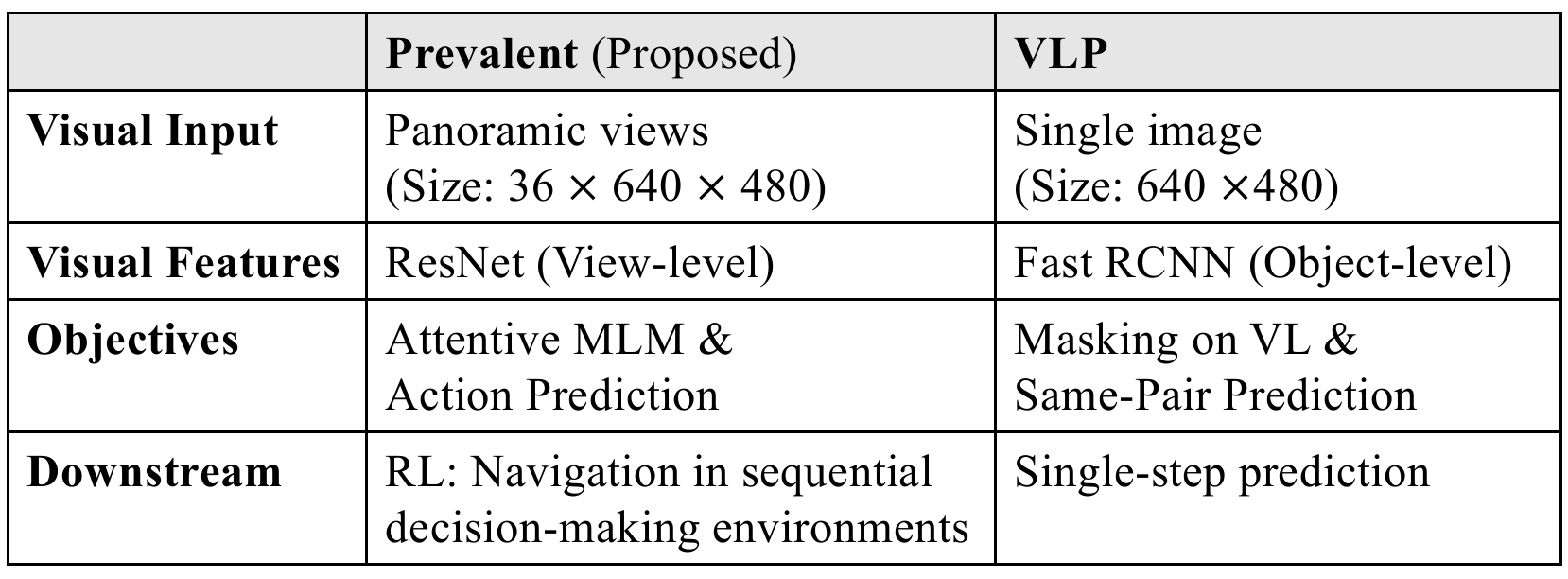}
	   \\
	   (a) \textsc{Press} &  (b) VLP
	  	\end{tabular}
	\end{minipage}
	\vspace{-3mm}
\caption{Comparison with related works.} 
	\label{tab:comparison}
\end{table}


%

\end{document}